\documentclass[a4paper,fleqn]{cas-dc}

\usepackage[numbers]{natbib}
\usepackage{rotating}
\usepackage{pdflscape}
\usepackage{subfig}
\usepackage{graphicx}
\usepackage{hyperref}
\def\tsc#1{\csdef{#1}{\textsc{\lowercase{#1}}\xspace}}
\tsc{WGM}
\tsc{QE}
\tsc{EP}
\tsc{PMS}
\tsc{BEC}
\tsc{DE}
\begin{document}
\let\WriteBookmarks\relax
\def\floatpagepagefraction{1}
\def\textpagefraction{.001}

\shorttitle{Taxonomy and Trends in Reinforcement Learning for Robotics and Control Systems: A Structured Review}

\shortauthors{Ter et~al.}

\title [mode = title]{Taxonomy and Trends in Reinforcement Learning for Robotics and Control Systems: A Structured Review}                      
%\tnotemark[1,2]

%\tnotetext[1]{This document is the results of the research
%   project funded by the National Science Foundation.}

%\tnotetext[2]{The second title footnote which is a longer text matter
%   to fill through the whole text width and overflow into
%   another line in the footnotes area of the first page.}

\author[1]{Kumater Ter}[
        %style=chinese
        ]
\ead{kumater.ter@afit.edu.ng}
\credit{Data curation}
\author[1]{Abolanle Adetifa}[%
   %role=Co-ordinator,
   %suffix=Jr,
   ]
\ead{adetifa@afit.edu.ng}
\credit{Data curation}
\author[1]{Daniel Udekwe}[
                        %type=editor,
                        %auid=000,bioid=1,
                        %prefix=Sir,
                        %role=Researcher,
                        orcid=0000-0003-1771-5320
                        ]

\cormark[1]


\ead{daudekwe@afit.edu.ng}

%\ead[url]{www.cvr.cc, cvr@sayahna.org}

\credit{Conceptualization of this study, Methodology, Software, Writing - Original draft preparation}

% \affiliation[1]{organization={FAMU-FSU College of Engineering, Florida State University},
%     addressline={2525 Pottsdamer St}, 
%     city={Tallahassee},
%     % citysep={}, % Uncomment if no comma needed between city and postcode
%     postcode={32310}, 
%     state={Florida},
%     country={USA}}

%\ead{rdonatus@afit.edu.ng}
%\ead[URL]{www.sayahna.org}

% \affiliation[2]{organization={Africa Centre of Excellence on Technology Enhanced Learning, National Open University},
%     % addressline={}, 
%     city={Lagos},
%     % citysep={}, % Uncomment if no comma needed between city and postcode
%     %postcode={695014}, 
%     %state={Trivandrum},
%     country={Nigeria}}

% \author%
% [1,3]
% {Rishi T.}
% \cormark[2]
% \fnmark[1,3]
% \ead{rishi@stmdocs.in}
% \ead[URL]{www.stmdocs.in}

\affiliation[1]{organization={Department of Aerospace of Engineering, Faculty of Air Engineering, Air Force Institute of Technology},
    %addressline={Mepukada}, 
    city={Kaduna},
    % citysep={}, % Uncomment if no comma needed between city and postcode
    %postcode={695571}, 
   % state={Trivandrum},
    country={Nigeria}}

\cortext[cor1]{Corresponding author}
%\cortext[cor2]{Principal corresponding author}

% \fntext[fn1]{This is the first author footnote. but is common to third
% \fntext[fn2]{Another author footnote, this is a very long footnote and
%   it should be a really long footnote. But this footnote is not yet
%   sufficiently long enough to make two lines of footnote text.}

% \nonumnote{This note has no numbers. In this work we demonstrate $a_b$
%   the formation Y\_1 of a new type of polariton on the interface
%   between a cuprous oxide slab and a polystyrene micro-sphere placed

\begin{abstract}
Reinforcement learning (RL) has become a foundational approach for enabling intelligent robotic behavior in dynamic and uncertain environments. This work presents an in-depth review of RL principles, advanced deep reinforcement learning (DRL) algorithms, and their integration into robotic and control systems. Beginning with the formalism of Markov Decision Processes (MDPs), the study outlines essential elements of the agent-environment interaction and explores core algorithmic strategies including actor-critic methods, value-based learning, and policy gradients. Emphasis is placed on modern DRL techniques such as DDPG, TD3, PPO, and SAC, which have shown promise in solving high-dimensional, continuous control tasks. A structured taxonomy is introduced to categorize RL applications across domains such as locomotion, manipulation, multi-agent coordination, and human-robot interaction, along with training methodologies and deployment readiness levels. The review synthesizes recent research efforts, highlighting technical trends, design patterns, and the growing maturity of RL in real-world robotics. Overall, this work aims to bridge theoretical advances with practical implementations, providing a consolidated perspective on the evolving role of RL in autonomous robotic systems.

\end{abstract}

% \includegraphics{figs/grabs.pdf}

% \begin{highlights}
% \item Research highlights item 1
% \item Research highlights item 2
% \item Research highlights item 3
% \end{highlights}

% Keywords
\begin{keywords}
Reinforcement Learning \sep Robotics \sep Control Systems \sep Deep Reinforcement Learning \sep Autonomous Systems
\end{keywords}
\maketitle

\section{Introduction}

Reinforcement Learning (RL), a subfield of machine learning, has emerged as a powerful paradigm for solving sequential decision-making problems \cite{shakya2023reinforcement, matsuo2022deep}. In RL, agents learn optimal behaviors through trial-and-error interactions with dynamic environments, guided by feedback in the form of rewards or penalties \cite{ladosz2022exploration}. This learning strategy is particularly well-suited for robotics and control systems, where agents must operate under uncertainty, interact with physical environments, and adapt to evolving tasks and conditions \cite{kaufmann2024survey}.

Traditional control methods, such as Proportional-Integral-Derivative (PID) controllers, Linear Quadratic Regulators (LQR), and Model Predictive Control (MPC), have long been the cornerstone of robotics \cite{guo2025deepseek}. These methods rely on accurate system models and often require manual tuning to perform well in complex, nonlinear, or changing environments \cite{elguea2023review}. Reinforcement learning, by contrast, offers the potential for model-free or partially model-based control strategies that can automatically adapt policies through direct interaction, even when the system dynamics are unknown or too complex to model precisely \cite{le2022deep}.

The confluence of RL with advances in deep learning, commonly referred to as Deep Reinforcement Learning (DRL), has enabled the application of these methods to high-dimensional and real-time control problems \cite{milani2024explainable}. This has led to significant breakthroughs in robotic locomotion, manipulation, autonomous navigation, and multi-agent coordination \cite{meyn2022control}. RL algorithms have shown promise not only in simulation but increasingly in real-world robotic systems, where they can outperform hand-engineered controllers and adapt to unstructured environments \cite{han2023survey}.

Despite its growing success, the deployment of RL in physical systems presents unique challenges, such as sample inefficiency, safety constraints, sim-to-real transfer gaps, and interpretability of learned behaviors \cite{han2023survey, almahamid2021reinforcement}. Addressing these issues requires a deep integration of RL methods with domain-specific knowledge from control theory, physics-based modeling, and hardware-aware system design.

This review aims to provide a comprehensive overview of the current landscape of reinforcement learning applications in robotics and control systems. We will:
\begin{itemize}
    \item Survey foundational concepts and core algorithms underpinning RL and DRL,
    \item Classify robotics based on taxonomy
    \item Highlight recent advances and real-world case studies in robotic locomotion, manipulation, and autonomous systems,
    \item Discuss practical challenges, benchmark environments, and open research directions.
\end{itemize}

By systematically organizing the field, this review serves both as an entry point for newcomers and a resource for researchers and practitioners aiming to push the frontiers of intelligent robotics.

\section{Background}

\subsection{Fundamentals of Reinforcement Learning}
\label{sec:fundamentals}

\begin{figure}
    \centering
    \includegraphics[width=0.5\textwidth]{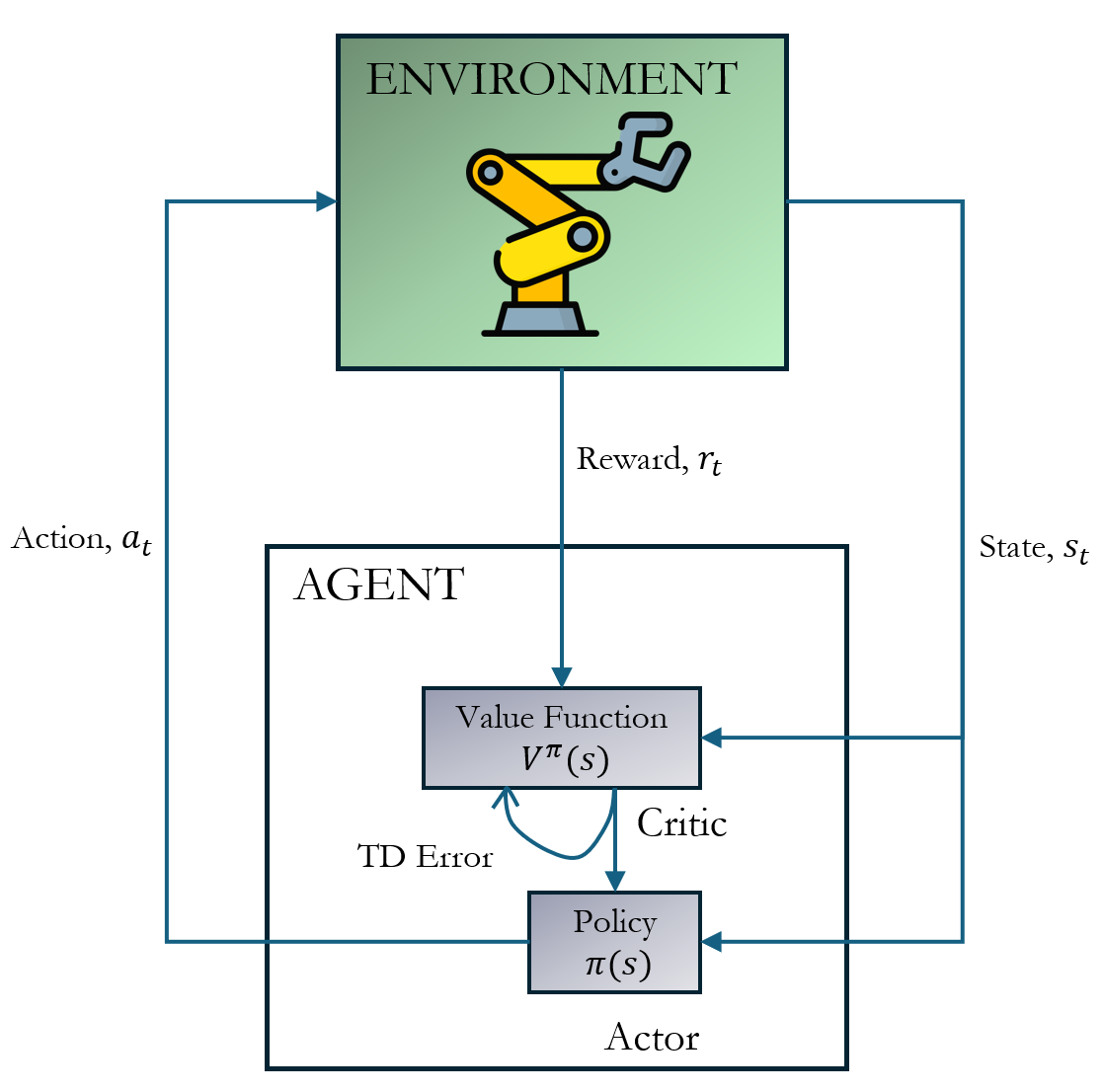}
    \caption{Illustration of the Actor-Critic reinforcement learning framework. The agent receives a state $s_t$ from the environment, selects an action $a_t$ via a policy $\pi(s)$, and receives a reward $r_t$. The critic evaluates the action using a value function $V^\pi(s)$.}
    \label{fig:rlsetup}
\end{figure}

Reinforcement Learning (RL) provides a framework in which agents learn to make sequential decisions by interacting with an environment and receiving feedback in the form of rewards \cite{okafor2021heuristic, okafor2021photovoltaic, okafor2021solar}. This process is commonly modeled as a Markov Decision Process (MDP), defined by the tuple $(\mathcal{S}, \mathcal{A}, \mathcal{P}, \mathcal{R}, \gamma)$ \cite{almahamid2021reinforcement, gronauer2022multi}, where:

\begin{itemize}
    \item $\mathcal{S}$ denotes the set of all possible environment states,
    \item $\mathcal{A}$ represents the set of actions the agent can take,
    \item $\mathcal{P}(s'|s,a)$ is the transition probability function, defining the likelihood of transitioning to state $s'$ after taking action $a$ in state $s$,
    \item $\mathcal{R}(s,a)$ is the reward function that quantifies the immediate gain from taking action $a$ in state $s$,
    \item $\gamma \in [0,1]$ is the discount factor, controlling the importance of future rewards relative to immediate ones.
\end{itemize}

As shown in Figure~\ref{fig:rlsetup}, an RL agent interacts with its environment over time. At each timestep $t$, it observes the current state $s_t$, selects an action $a_t$ based on its policy $\pi(s)$, and receives a scalar reward $r_t$ from the environment. The environment then transitions to a new state $s_{t+1}$, continuing the cycle.

Modern RL methods often employ a division between two core modules: the actor and the critic. The actor is responsible for selecting actions, whereas the critic estimates the value of those actions using a value function, such as $V^\pi(s)$ for state values or $Q^\pi(s, a)$ for state-action pairs. The critic evaluates the actor's choices and provides feedback in the form of temporal-difference (TD) error, which the actor uses to improve its policy. This actor-critic architecture forms the basis of many deep RL algorithms, inc

The goal of the agent is to learn a policy $\pi(a|s)$ that maximizes the expected cumulative discounted return \cite{prudencio2023survey, adetifa2023deep}:

\begin{equation}
    J(\pi) = \mathbb{E}_{\pi} \left[ \sum_{t=0}^{\infty} \gamma^t r_t \right]
\end{equation}

The value function $V^{\pi}(s)$ estimates the expected return from state $s$ under policy $\pi$ \cite{szepesvari2022algorithms, udekwe2024comparing}:

\begin{equation}
    V^{\pi}(s) = \mathbb{E}_{\pi} \left[ \sum_{t=0}^{\infty} \gamma^t r_t \mid s_0 = s \right]
\end{equation}

Similarly, the action-value function $Q^{\pi}(s, a)$ is \cite{laskin2021urlb, udekwe2025evaluating}:

\begin{equation}
    Q^{\pi}(s,a) = \mathbb{E}_{\pi} \left[ \sum_{t=0}^{\infty} \gamma^t r_t \mid s_0 = s, a_0 = a \right]
\end{equation}

Learning optimal policies involves estimating the optimal value functions \cite{laskin2021urlb}:

\begin{align}
    V^*(s) &= \max_{\pi} V^{\pi}(s) \\
    Q^*(s,a) &= \max_{\pi} Q^{\pi}(s,a)
\end{align}

Algorithms such as Q-learning update estimates of $Q(s,a)$ using the Bellman equation \cite{adams2022survey}:

\begin{equation}
    Q(s,a) \leftarrow Q(s,a) + \alpha \left[ r + \gamma \max_{a'} Q(s', a') - Q(s,a) \right]
\end{equation}

where $\alpha$ is the learning rate.

\subsection{Traditional Control Methods in Robotics}

Conventional control in robotics is often based on analytical modeling and linear system theory. Examples include:

\begin{itemize}
    \item PID control:\ Computes control inputs as a combination of proportional, integral, and derivative terms \cite{borase2021review}:
    \begin{equation}
        u(t) = K_p e(t) + K_i \int_{0}^{t} e(\tau) d\tau + K_d \frac{de(t)}{dt}
    \end{equation}
    where $e(t)$ is the error signal and $K_p$, $K_i$, $K_d$ are tuning parameters.

    \item Linear Quadratic Regulator (LQR): Minimizes a quadratic cost function for linear systems \cite{chacko2023lqr}:
    \begin{equation}
        J = \int_0^{\infty} \left( x^T Q x + u^T R u \right) dt
    \end{equation}
    where $x$ is the system state and $u$ is the control input.

    \item Model Predictive Control (MPC): Solves an optimization problem over a finite prediction horizon using a model of the system \cite{schwenzer2021review}:
    \begin{equation}
        \min_{u_{0:H}} \sum_{t=0}^{H} \left( x_t^T Q x_t + u_t^T R u_t \right) \quad \text{s.t. dynamics and constraints}
    \end{equation}
\end{itemize}

While these methods are effective for known dynamics and structured environments, they often require precise modeling and extensive manual tuning, which limits their adaptability.

\subsection{Deep Reinforcement Learning (DRL)}

Deep Reinforcement Learning (DRL) integrates the decision-making framework of Reinforcement Learning with the function approximation power of deep neural networks \cite{zhao2021dynamic}. This combination allows agents to operate in high-dimensional, continuous, or partially observed environments, making DRL particularly suited to complex robotics and control tasks such as locomotion, manipulation, and navigation \cite{zhao2021dynamic}.

At the core of DRL is the idea of using a neural network to approximate value functions, policies, or models of the environment. In control settings, DRL often deals with continuous state and action spaces, where traditional tabular RL methods become intractable.
\\
\subsubsection*{1. Deep Q-Network (DQN)}

The DQN algorithm approximates the Q-value function $Q(s, a; \theta)$ using a neural network with parameters $\theta$. The network is trained to minimize the Bellman error \cite{zhao2021dynamic}:

\begin{equation}
    L(\theta) = \mathbb{E}_{(s, a, r, s')} \left[ \left( y^{DQN} - Q(s,a;\theta) \right)^2 \right]
\end{equation}

where the target is defined as:

\begin{equation}
    y^{DQN} = r + \gamma \max_{a'} Q(s', a'; \theta^-)
\end{equation}

Here, $\theta^-$ denotes the parameters of a target network, which are periodically updated to stabilize learning. DQN is suitable for discrete action spaces and was originally applied to Atari games using pixel inputs.
\\
\subsubsection*{2. Deep Deterministic Policy Gradient (DDPG)}

For continuous action spaces, DDPG combines Q-learning with deterministic policy gradients. It maintains two networks \cite{bouhamed2020autonomous}:

\begin{itemize}
    \item A critic network $Q(s,a;\theta^Q)$ to estimate action-values.
    \item An actor network $\mu(s;\theta^\mu)$ to output deterministic actions.
\end{itemize}

The critic is trained using:

\begin{equation}
    y^{DDPG} = r + \gamma Q(s', \mu(s'; \theta^\mu); \theta^Q)
\end{equation}

and the actor is updated via the deterministic policy gradient:

\begin{equation}
    \nabla_{\theta^\mu} J \approx \mathbb{E}_s \left[ \nabla_a Q(s, a; \theta^Q) \vert_{a = \mu(s)} \nabla_{\theta^\mu} \mu(s) \right]
\end{equation}

DDPG is widely used in robotic control tasks due to its ability to handle continuous action spaces.
\\
\subsubsection*{3. Twin Delayed DDPG (TD3)}

TD3 improves upon DDPG by addressing the problem of overestimation bias and training instability. It introduces \cite{zhang2020td3}:

\begin{itemize}
    \item Clipped Double Q-learning: Two critics are used to compute the minimum Q-value.
    \item Delayed policy updates: The actor is updated less frequently than the critic.
    \item Target policy smoothing: Small noise is added to the target action to reduce variance.
\end{itemize}

\subsubsection*{4. Soft Actor-Critic (SAC)}

SAC is an off-policy algorithm that maximizes a trade-off between expected return and policy entropy \cite{de2021soft}:

\begin{equation}
    J(\pi) = \sum_{t=0}^{\infty} \mathbb{E}_{(s_t,a_t) \sim \pi} \left[ r(s_t, a_t) + \alpha \mathcal{H}(\pi(\cdot|s_t)) \right]
\end{equation}

where $\mathcal{H}(\pi)$ is the entropy of the stochastic policy and $\alpha$ is a temperature parameter. By encouraging stochasticity, SAC improves exploration and robustness in continuous control tasks.
\\
\subsubsection*{5. Proximal Policy Optimization (PPO)}

PPO is a policy gradient method that balances exploration and stability using a clipped surrogate objective \cite{huang2022a2c}:

\begin{equation}
\begin{aligned}
    L^{CLIP}(\theta) = \mathbb{E}_t \big[ \min \big( r_t(\theta) \hat{A}_t, \\
    \text{clip}(r_t(\theta), 1 - \epsilon, 1 + \epsilon) \hat{A}_t \big) \big]
\end{aligned}
\end{equation}

where $r_t(\theta) = \frac{\pi_\theta(a_t|s_t)}{\pi_{\theta_{old}}(a_t|s_t)}$ is the probability ratio and $\hat{A}_t$ is the estimated advantage. PPO has become a standard baseline for continuous and discrete control due to its simplicity and effectiveness.
\\
\subsubsection*{6. Asynchronous Advantage Actor-Critic (A3C)}

A3C utilizes multiple agents in parallel environments to stabilize training. Each worker computes gradients for an actor and critic \cite{chen2020intelligent}:

\begin{itemize}
    \item Actor: updates policy parameters using the advantage estimate.
    \item Critic: estimates the value function for advantage calculation.
\end{itemize}

The parallelism reduces variance and accelerates convergence, making A3C suitable for environments with long episodes and delayed rewards.
\\
\subsubsection*{7. Other Advanced Algorithms}

Recent developments include:

\begin{itemize}
    \item Distributional RL (e.g., C51, QR-DQN): Models the full return distribution instead of expected value \cite{fathan2021deep, zwetsloot2025evaluating}.
    \item Hierarchical RL (e.g., HIRO): Learns policies at multiple temporal scales for better abstraction and planning \cite{zhang2021hierarchical}.
    \item Meta-RL (e.g., MAML, PEARL): Trains agents that adapt rapidly to new tasks using limited experience \cite{fu2023maml2, zhu2024pearl}.
\end{itemize}

These DRL algorithms provide a flexible and powerful toolkit for robotics, enabling learning from visual input, generalization across tasks, and control of complex, high-dimensional systems.

\section{Taxonomy of Reinforcement Learning Applications in Robotics}
\label{sec:taxonomy}

\begin{figure*}
    \centering
    \includegraphics[width=\textwidth]{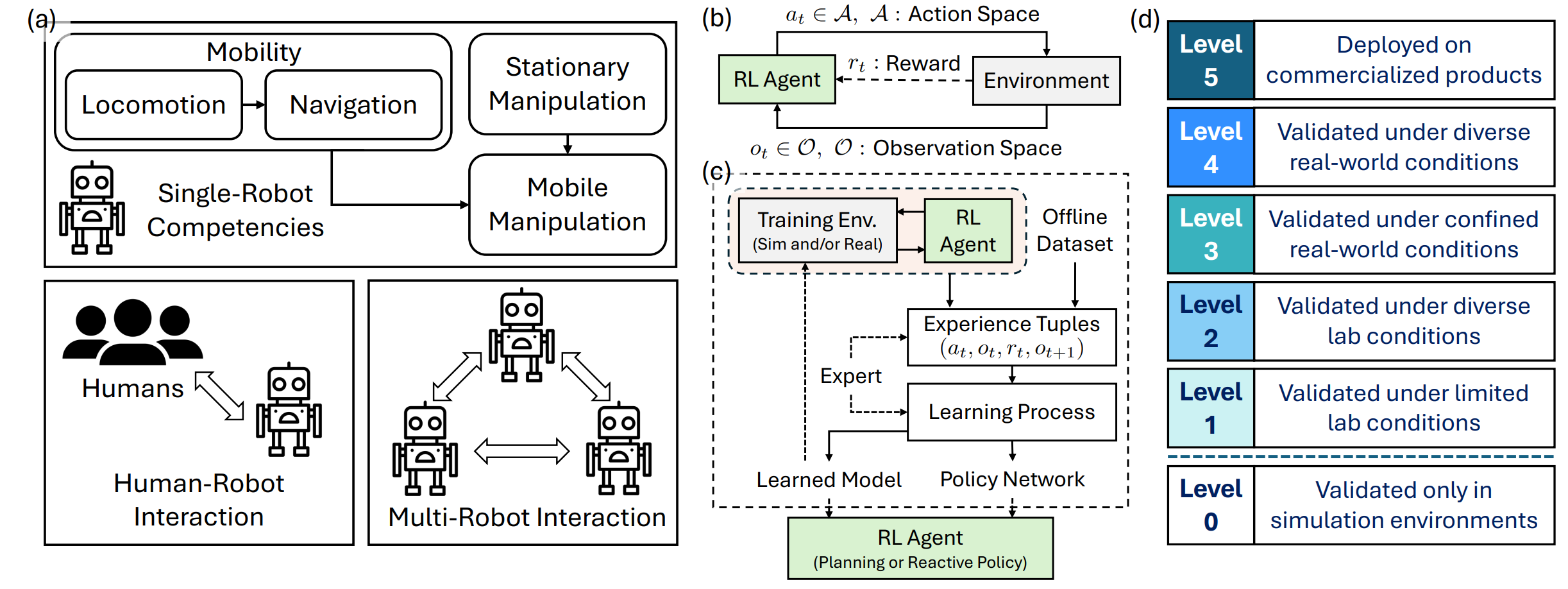}
    \caption{Taxonomy of RL Applications in Robotics and Control Systems. Components (a)--(d) depict robot task categories, training structure, learning pipeline, and deployment readiness levels \cite{tang2025deep}}
    \label{fig:taxonomy}
\end{figure*}

The taxonomy illustrated in Figure~\ref{fig:taxonomy} provides a structured view of how reinforcement learning (RL) is applied in various robotics contexts. It synthesizes prior efforts into four primary components: robotic skill domains and interaction structures (a), standard RL formalism (b), learning pipeline architectures (c), and a framework for assessing deployment readiness (d). This classification helps clarify how RL integrates into robotic systems of varying complexity and deployment stages.

\subsection*{(a) Competencies and Interaction Modalities in Robotic Systems}
The first component of the taxonomy categorizes RL applications based on robot capabilities and interaction dynamics \cite{garaffa2021reinforcement}. In the domain of single-robot operation, key competencies are grouped into locomotion, navigation, stationary manipulation, and mobile manipulation. Locomotion refers to the robot's fundamental ability to move, typically through legs, wheels, or tracks \cite{tang2025deep}. Navigation builds upon locomotion by introducing goal-directed movement in structured or unstructured environments \cite{singh2022reinforcement}. In parallel, manipulation tasks may occur in stationary settings (e.g., robotic arms in manufacturing) or in mobile contexts, where navigation and manipulation are integrated \cite{ibarz2021train}.

Beyond single-robot systems, the taxonomy recognizes more complex interactive settings \cite{tang2025deep}. Human-robot interaction (HRI) encompasses tasks where humans provide feedback, demonstrations, or shared decision-making input to guide RL behavior \cite{singh2022reinforcement}. Multi-robot interaction introduces coordination across multiple autonomous agents, requiring communication protocols, task decomposition, and policy synchronization. These two domains emphasize the growing complexity and social integration of RL-driven robotic systems \cite{garaffa2021reinforcement}.

\subsection*{(b) The Reinforcement Learning Loop: Agent and Environment}
At the core of RL applications lies the classical agent-environment interaction cycle \cite{ibarz2021train}. An RL agent receives observations from the environment and responds with an action, which leads to a new observation and a scalar reward signal \cite{korber2021comparing}. This cycle allows the agent to iteratively improve its policy through trial and error, with the goal of maximizing cumulative rewards over time. The environment may be physical or simulated, and the agent's policy can be deterministic or stochastic \cite{hua2021learning}. This formulation underpins nearly all reinforcement learning algorithms, whether value-based, policy-gradient, or actor-critic in nature \cite{zhu2021deep}.

\subsection*{(c) Training Pipelines and Learning Architectures}
The third component delves into how RL agents are trained. This includes a mix of online training in simulation or physical environments, and offline learning from logged data \cite{yang2020data}. In many scenarios, an RL agent collects experience tuples—consisting of actions, observations, rewards, and next states—which are then used to update the policy or value functions \cite{oliff2020reinforcement}. The learning process may incorporate demonstrations from expert agents or humans to improve sample efficiency and reduce unsafe exploration, especially in robotics applications with high-stakes outcomes \cite{wang2020mobile}. Eventually, the result of this pipeline is a trained policy network or model that can be deployed for real-time decision making, either reactively or as part of a planning system \cite{oliff2020reinforcement}.

\subsection*{(d) Readiness Levels for Deployment in Real-World Settings}
Finally, the taxonomy introduces a maturity model that categorizes RL systems based on their level of real-world deployment \cite{yang2020data}. At level 0, policies are evaluated only in simulation, with no exposure to real-world uncertainty \cite{oliff2020reinforcement}. Level 1 introduces limited laboratory validation, while level 2 expands testing across diverse lab setups \cite{joshi2020robotic}. Level 3 marks the transition to real-world environments, albeit in controlled conditions \cite{yang2020data}. Level 4 systems demonstrate robustness under more diverse and dynamic real-world scenarios \cite{zhang2022reinforcement}. Ultimately, level 5 indicates deployment in commercial or industrial products. This readiness scale is useful for evaluating how far RL-based robotic systems have progressed from conceptual prototypes to robust, deployed technologies.

\section{Key Applications}

Reinforcement Learning has enabled significant progress in various robotic domains, particularly in areas requiring adaptability, decision-making under uncertainty, and continuous learning. This section highlights four major application areas where RL has demonstrated promising or state-of-the-art performance: locomotion, manipulation, autonomous navigation, and industrial automation.

\subsection{Locomotion}

RL has revolutionized the field of robotic locomotion, particularly for legged and non-wheeled robots. These systems operate in high-dimensional, continuous action spaces with nonlinear dynamics, making them suitable for RL-based control.

\begin{itemize}
    \item Bipedal and Quadrupedal Walking: Policies trained using Proximal Policy Optimization (PPO) and Soft Actor-Critic (SAC) have achieved dynamic walking and running behaviors on simulated and physical platforms such as MIT Cheetah, ANYmal, and Cassie \cite{weng2021natural}.
    \item Stability and Adaptability: RL policies can learn to adapt to uneven terrain, external perturbations, or joint failures without explicit reprogramming \cite{jin2020stability}.
    \item Hardware Deployment: Sim-to-real transfer is accomplished using domain randomization and curriculum learning to bridge the reality gap \cite{zhao2020sim}.
\end{itemize}

\subsection{Manipulation and Grasping}

Robotic manipulation tasks such as pick-and-place, in-hand reorientation, and object assembly are complex due to contact dynamics, partial observability, and precision requirements. RL enables data-driven learning of dexterous manipulation policies.

\begin{itemize}
    \item End-to-End Control: Deep RL is used to map raw images or proprioceptive signals directly to joint torques or velocities \cite{zhang2021end, ajayi2025integrating}.
    \item Contact-Rich Tasks: Algorithms like DDPG and SAC have been used to train multi-fingered robotic hands (e.g., ShadowHand) for tasks like object flipping and tool use \cite{spector2020deep}.
    \item Learning from Demonstrations (LfD): Combining RL with imitation learning improves sample efficiency and safety during training \cite{ibarz2021train}.
\end{itemize}

\subsection{Autonomous Navigation}

Mobile robots operating in indoor or outdoor environments must learn to navigate, avoid obstacles, and plan efficient paths in real time. RL contributes to decision-making in both global path planning and local control.

\begin{itemize}
    \item Dynamic Environments: RL agents learn to react to moving obstacles and changing goals using visual or LiDAR-based inputs \cite{zhang2021end, udekwe2022development}.
    \item End-to-End Visual Navigation: CNNs trained with RL can directly output control commands from camera feeds in cluttered environments \cite{jin2020stability}.
    \item Multi-Modal Fusion: RL architectures fuse GPS, IMU, vision, and range sensors to learn robust navigation strategies \cite{weng2021natural}.
\end{itemize}

\subsection{Industrial Automation}

In industrial settings, RL is increasingly used for flexible automation and adaptive control in dynamic production environments.

\begin{itemize}
    \item Robotic Assembly: RL is used for force-sensitive insertion, gear alignment, and part manipulation with real-time feedback \cite{ahn2023robotic, donatus2025multi}.
    \item Adaptive Quality Control: Inspection robots trained with RL can adapt their scanning behavior based on defect likelihood particularly with virtual reality \cite{viharos2021reinforcement, udekwe2025human, udekwe2025virtual}.
    \item Process Optimization: RL controls temperature, pressure, or conveyor belt speed to optimize energy efficiency and throughput \cite{petsagkourakis2020reinforcement}.
\end{itemize}

\begin{sidewaystable*}
\centering
\caption{Review of Reinforcement Learning Applications in Robotics and Control Systems}
\begin{tabular}{p{2.5cm}p{5.5cm}p{7cm}p{4.5cm}}
\hline
\textbf{Study} & \textbf{Objective} & \textbf{Technique} & \textbf{Domain}  \\
\hline
\cite{luo2025precise} & Dexterous robotic manipulation with real-world RL & Human-in-the-loop RL, RLPD, vision-based policy, dual-arm tasks & Industrial robot manipulation \\
\\
\cite{bai2024digirl} & Device control through GUIs using VLMs and RL & Offline + online RL, Advantage-weighted regression, AutoEval, curriculum learning & Digital agents / GUI control \\
\\
\cite{meng2024online} & Efficient energy management in microgrids under uncertainty & Online RL, SARSA, WPESS model optimization & Smart grids / energy control \\
\\
\cite{chu2025adaptive} & Robust underwater vehicle docking & Adaptive Reward Shaped PPO (ARSPPO), MDP modeling, simulation + experiments & Autonomous underwater vehicles (AUV) \\
\\
\cite{liu2024deep} & Efficient mobile robot path planning with DRL & Parallel SAC (PSAC), hybrid A* path correction, embedded testing & Mobile robot navigation \\
\\
\cite{xu2024reinforcement} & Efficient multi-robot scheduling in smart factories & Q-learning with event-triggered updates, decentralized coordination & Industrial scheduling and manufacturing \\
\\
\cite{wu2024multi} & Safe navigation under uncertain terrain conditions & Hierarchical RL, distributional RL for uncertainty modeling & Autonomous ground vehicles \\
\\
\cite{wang2024research} & Dynamic quadruped locomotion with learning-based torque control & Actor-Critic with torque-to-command mapping, residual control & Legged robotics \\
\\
\cite{WANG2023112778} & Learning-based task planning in human-robot collaboration & DRL + symbolic task representation, reward shaping & Collaborative robotics \\
\\
\cite{fu2023ed} & Vision-based robotic bin-picking with high success rate & End-to-end DRL with 3D object localization and policy fusion & Industrial robotics / object manipulation \\
\\
\cite{wu2023deep} & Secure control of cyber–physical systems (CPS) under actuator false data injection attacks & Lyapunov-based Soft Actor–Critic (SAC), proving convergence and exponential stability & Cyber–physical systems, secure robot arm control, satellite attitude control \\

\hline
\end{tabular}
\label{tab:rl_review_sideways}
\end{sidewaystable*}

\begin{sidewaystable*}
\centering
\caption{Review of Reinforcement Learning Applications in Robotics and Control Systems}
\begin{tabular}{p{2.5cm}p{5.5cm}p{7cm}p{4.5cm}}
\hline
\textbf{Study} & \textbf{Objective} & \textbf{Technique} & \textbf{Domain} \\
\hline
\cite{zhou2024indoor} & Multi-agent coordination for UAV swarm inspection tasks & Graph neural network-enhanced multi-agent RL (GNN-MARL) & Aerial robotics, multi-UAV collaboration \\
\\
\cite{zhang2024multi} & Adaptive control in robotic arm tasks with minimal supervision & Self-supervised reinforcement learning, sparse reward optimization & Industrial robot arms, adaptive manipulation \\
\\
\cite{yang2024human} & Real-time control for autonomous underwater vehicles & Hierarchical DDPG, safety-constrained RL, environment modeling & Marine robotics, underwater exploration \\
\\
\cite{wang2024distrl} & Energy-efficient HVAC control using meta-RL & Model-free RL, meta-policy optimization & Smart buildings, HVAC systems \\
\\
\cite{al2024reinforcement} & Review of RL for intelligent HVAC systems with focus on generalization & Survey of model-free RL (e.g., SAC, PPO), meta-RL for better adaptability & Building energy systems, HVAC control \\
\\
\cite{silvestri2024real} & Real-world HVAC control using DRL for energy efficiency & Soft Actor-Critic (SAC), model calibration with real building data & Thermal building systems, DRL deployment \\
\\
\cite{tang2025deep} & Survey of real-world successes of DRL in robotics & Comprehensive survey; analysis of policy learning, deployment techniques, and challenges & Robotics (general), multi-domain DRL applications \\
\\
\cite{xu2024autonomous} & Autonomous navigation for unmanned vehicles using DRL & Deep Deterministic Policy Gradient (DDPG) with Ackermann steering kinematics & Autonomous driving, mobile robotics \\
\\
\cite{yin2024autonomous} & Robust quadrotor control in complex environments using DRL & TD3 algorithm with environmental awareness for path tracking & Aerial robotics, quadrotor control \\
\\
\cite{esteso2023reinforcement} & Safe and optimal energy consumption in robot-assisted logistics & Safe RL using Constrained Policy Optimization (CPO) for trajectory planning & Warehouse robotics, logistics automation \\
\\
\cite{zhu2023intelligent} & Multi-agent collaboration for drone fleets using RL & Multi-agent reinforcement learning (MARL) with reward shaping & Multi-drone coordination, aerial surveillance \\
\hline
\end{tabular}
\label{tab:rl_review_additional}
\end{sidewaystable*}

\section{Challenges in Real-World Deployment}

Despite significant advances in simulation and algorithmic development, deploying Reinforcement Learning (RL) in real-world robotic systems remains a non-trivial endeavor. Practical applications face challenges related to sample efficiency, safety, generalization, and system constraints. This section outlines the primary obstacles that limit the scalability and robustness of RL in physical robotics.

\subsection{Sample Inefficiency}

Most RL algorithms require millions of environment interactions to converge to optimal or even usable policies. In the real world, such interactions are time-consuming, wear out hardware, and can be unsafe.

\begin{itemize}
    \item Data-Hungry Algorithms: Algorithms like DDPG, SAC, and PPO need large volumes of data, which is often impractical outside simulation \cite{li2022reinforcement}.
    \item Solutions: Off-policy learning, model-based RL, and leveraging prior knowledge (e.g., demonstrations) are common approaches to reduce sample complexity \cite{wang2020reinforcement}.
\end{itemize}

\subsection{Safety and Stability}

Real-world robots interact with fragile environments, expensive equipment, and even humans. Unsafe actions during learning can cause irreversible damage.

\begin{itemize}
    \item Exploration Risk: Naïve exploration strategies can result in collisions, falls, or unsafe torque outputs \cite{nguyen2020deep}.
    \item Mitigations: Constrained RL, safety filters (e.g., Control Barrier Functions), and human-in-the-loop frameworks are being developed to enforce operational limits \cite{wong2023deep}.
\end{itemize}

\subsection{Sim-to-Real Transfer Gap}

Policies trained in simulation often fail to generalize to real hardware due to discrepancies in dynamics, sensors, friction, and latency.

\begin{itemize}
    \item Reality Gap: Even small modeling errors or delays can cause learned policies to become unstable when deployed on real systems \cite{du2021survey}.
    \item Bridging Techniques: Domain randomization, system identification, and adaptive residual policies are used to narrow the sim-to-real gap \cite{hutsebaut2022hierarchical}.
\end{itemize}

\subsection{Sparse and Delayed Rewards}

In many robotics tasks, meaningful feedback is only available after completing a complex sequence of actions, making it difficult for agents to assign credit correctly.

\begin{itemize}
    \item Delayed Feedback: Tasks like assembly or manipulation often involve long horizons with infrequent rewards \cite{zhou2023multi}.
    \item Solutions: Reward shaping, curriculum learning, and hierarchical reinforcement learning can guide agents more effectively \cite{ilahi2021challenges}.
\end{itemize}

\subsection{Generalization and Transferability}

Policies often overfit to specific environments, robot configurations, or training conditions and fail when exposed to unseen variations.

\begin{itemize}
    \item Domain Sensitivity: Learned behaviors might degrade when dynamics change due to wear, load, or environment\cite{zhu2020ingredients}.
    \item Ongoing Research: Meta-RL, transfer learning, and robust RL aim to develop policies that generalize across tasks and domains \cite{parker2022automated}.
\end{itemize}

\subsection{Computational Constraints and Real-Time Execution}

Robots require policies to operate within tight latency bounds. Complex RL models may not meet the runtime demands of embedded control systems.

\begin{itemize}
    \item Latency Limits: Deep neural networks might not run at high-enough frequency for fast control loops \cite{levine2020offline}.
    \item Optimization: Policy distillation, network pruning, and hardware-aware policy compression techniques are increasingly adopted to reduce computation time \cite{gronauer2022multi}.
\end{itemize}

\subsection{Interpretability and Debugging}

Unlike traditional controllers with analytically derived behavior, RL policies are often treated as black boxes, making it difficult to interpret or diagnose failures.

\begin{itemize}
    \item Opaque Policies: Neural policies lack explainability, making failure analysis and trustworthiness difficult \cite{wong2023deep}.
    \item Research Efforts: Visualization tools, saliency maps, and explainable RL models are being explored to enhance transparency \cite{nguyen2020deep}.
\end{itemize}

\section{Future Directions and Conclusion}

Reinforcement Learning (RL) continues to transform robotics by enabling machines to learn from interaction, adapt to uncertainty, and autonomously acquire complex behaviors. While recent progress has demonstrated the feasibility of using RL in both simulated and real-world robotic systems, many foundational challenges remain. This final section outlines several promising research directions that could further expand the scope, reliability, and impact of RL in robotics, followed by a concluding summary.

\subsubsection*{Human-in-the-Loop Reinforcement Learning}

Incorporating human feedback into the RL loop can improve safety, accelerate learning, and align robotic behavior with user preferences. Techniques such as reward shaping via demonstration, inverse reinforcement learning (IRL), and preference-based learning are key enablers of interactive robot training.

\subsubsection*{Lifelong and Continual Learning}

Robots deployed in open-world settings must adapt to non-stationary environments, evolving tasks, and long-term operational changes. Lifelong RL algorithms aim to retain and reuse prior knowledge without catastrophic forgetting, enabling continual skill refinement and task expansion.

\subsubsection*{Causal Reinforcement Learning}

Integrating causal reasoning into RL frameworks can lead to better generalization, safer exploration, and more data-efficient learning. Causal RL models can differentiate correlation from causation, allowing agents to reason about interventions and counterfactuals during decision-making.

\subsubsection*{Hybrid RL and Classical Control}

Combining RL with traditional control methods like PID or MPC can yield more interpretable, stable, and sample-efficient systems. Hybrid frameworks can leverage the structure and guarantees of classical controllers while using RL to adaptively tune parameters or handle unmodeled dynamics.

\subsubsection*{Policy Generalization and Transfer Learning}

Developing agents that generalize across tasks, robots, and environments remains a grand challenge. Meta-learning, domain adaptation, and modular policy architectures are promising approaches to transfer skills across variations with minimal fine-tuning.

\subsubsection*{Scalable Simulation and Sim-to-Real Pipelines}

Large-scale simulation platforms (e.g., Isaac Gym, Mujoco, Habitat) enable faster experimentation but require robust sim-to-real techniques for deployment. Future research must improve the fidelity of simulators and devise adaptive transfer pipelines to minimize reliance on handcrafted calibration.

\subsubsection*{Explainability and Trustworthy RL}

As RL-based systems operate in critical domains, their behavior must be transparent and verifiable. Research on explainable RL, interpretable neural policies, and formal verification is crucial to increase trust and adoption in real-world deployments.

\section{CONCLUSION} \label{conclusion}

This review has explored the growing role of reinforcement learning in robotics and control systems, highlighting its applications across locomotion, manipulation, navigation, and multi-agent systems. We presented a taxonomy of RL-based robotic tasks, discussed representative algorithms and architectures, and outlined the practical challenges faced in real-world deployment, including sample inefficiency, safety, and generalization gaps.

The field is advancing rapidly, fueled by breakthroughs in deep learning, simulation tools, and hardware platforms. However, widespread adoption of RL in industry and mission-critical robotics still requires solving foundational issues related to robustness, interpretability, and efficiency.

Looking forward, the convergence of RL with causality, control theory, human-in-the-loop design, and lifelong learning presents a rich landscape for future exploration. As these areas mature, reinforcement learning holds the potential to create truly autonomous, adaptive, and intelligent robotic systems capable of operating seamlessly in complex, real-world environments.

% \appendix
% \section{My Appendix}
% Appendix sections are coded under \verb+\appendix+.

% \verb+\printcredits+ command is used after appendix sections to list 
% author credit taxonomy contribution roles tagged using \verb+\credit+ 
% in frontmatter.

\printcredits

\bibliographystyle{cas-model2-names}

\bibliography{cas-refs}


% Author biography without author photo.
% Author biography. Author biography. Author biography.
% Author biography. Author biography. Author biography.
% Author biography. Author biography. Author biography.
% Author biography. Author biography. Author biography.
% Author biography. Author biography. Author biography.
% Author biography. Author biography. Author biography.
% Author biography. Author biography. Author biography.
% Author biography. Author biography. Author biography.
% Author biography. Author biography. Author biography.

% \bio{figs/pic1}
% Author biography with author photo.
% Author biography. Author biography. Author biography.
% Author biography. Author biography. Author biography.
% Author biography. Author biography. Author biography.
% Author biography. Author biography. Author biography.
% Author biography. Author biography. Author biography.
% Author biography. Author biography. Author biography.
% Author biography. Author biography. Author biography.
% Author biography. Author biography. Author biography.
% Author biography. Author biography. Author biography.

% \bio{figs/pic1}
% Author biography with author photo.
% Author biography. Author biography. Author biography.
% Author biography. Author biography. Author biography.
% Author biography. Author biography. Author biography.
% Author biography. Author biography. Author biography.

\end{document}